# The influence of the focus in the camera calibration process

Carlos Ricolfe-Viala

**Abstract**—— Camera calibration is a crucial step in robotics and computer vision. Accurate camera parameters are necessary to achieve robust applications. Nowadays, camera calibration process consists of adjusting a set of data to a pin-hole model, assuming that with a reprojection error close to cero, camera parameters are correct. Since all camera parameters are unknown, computed results are considered true. However, the pin-hole model does not represent the camera behavior accurately if the focus is considered. Real cameras change the focal length slightly to obtain sharp objects in the image and this feature skews the calibration result if a unique pin-hole model is computed with a constant focal length.

In this paper, a deep analysis of the camera calibration process is done to detect and strengthen its weaknesses. The camera is mounted in a robot arm to known extrinsic camera parameters with accuracy and to be able to compare computed results with the true ones. Based on the bias that exist between computed results and the true ones, a modification of the widely accepted camera calibration method using images of a planar template is presented. A pin-hole model with distance dependent focal length is proposed to improve the calibration process substantially.

## 1 Introduction

CAMERA calibration is an essential issue in robotics and computer vision because it establishes the geometric relation between 2D image coordinates and 3D world coordinates [1, 2, 3, 4]. Many published papers explain how to obtain the correct mapping between 3D space and the 2D camera plane. Most of the work is based on the pinhole camera model using 3D [5, 6], 2D [7, 8] or 1D [9, 10, 11, 12] templates or doing self-calibration [13,14]. Photogrammetric methods use precise coordinates of calibration points in 3D space, arranged in a predesigned 3D, 2D or 1D template. Self-calibration methods assume that several images of a rigid scene with fixed camera parameters is enough to compute them.

Most existing methods propose a nonlinear minimization step that computes the correct camera parameters by iteratively minimizing the difference between the detected control points in images and their computed projections in the image plane. This difference is called the reprojection error [15]. A closed-form linear transformation solution initializes the nonlinear minimization step. The closed-form solution obtains an approximation of all linear camera parameters and the nonlinear minimization improves this approximation, together with nonlinear camera parameters such as lens distortion. The nonlinear minimization process ends when the reprojection error is close to zero. If the reprojection error is zero, the difference between the detected points in the image and the control points projected by the computed model is also zero. In consequence, it is assumed that the computed model is correct.

The problem arises when the distance from the camera to the calibration template vary in each image and the camera focus changes the lens position to obtain a focused image. In this case, images with different focal length are used to compute a unique model with a constant focal length. The reprojection error verifies that the computed model is satisfied with the input data, but it does not confirm that the computed model represents the real camera and the true camera location when the calibration images were taken.

If the camera-template distance varies in each image and a unique model with a constant focal lenght is computed, defocused images should be used. Several authors propose different methods to detect template control points in defocused images accurately [16 - 22]. This is out of the scope of this paper.

This aim of this paper is to analyse the influence of the camera focus in the results of the camera calibration process and to define a set of rules that will help to improve the outcomes when a camera is calibrated using an autofocus lens. The camera is a system in which intrinsic and extrinsic camera parameters are tightly coupled. When parameters are interdependent, a nonlinear minimization process that simultaneously computes all parameters together may not be the best choice. It is assumed that randomly acquired images reduce computed camera parameters bias but this is not always true as pointed out by Hu and Kantor [16]. Ryusuke and Y. Yasushi [17] proposed a separate calibration method for intrinsic camera parameters, but special equipment and a controlled environment are required. Alturki [18] and Lu [19] compute the principal point singularly by finding orthogonal projections of the camera optical axis on the image plane.

This paper proposes a calibration process of a camera with autofocus lenses modifiying a well-known camera

• *Carlos Ricolfe-Viala is with the Institute of Industrial Informatics and Automatic Control in Polythechic University of Valencia (UPV), Spain +34 963 877007; e-mail: cricolfe@upv.es).*

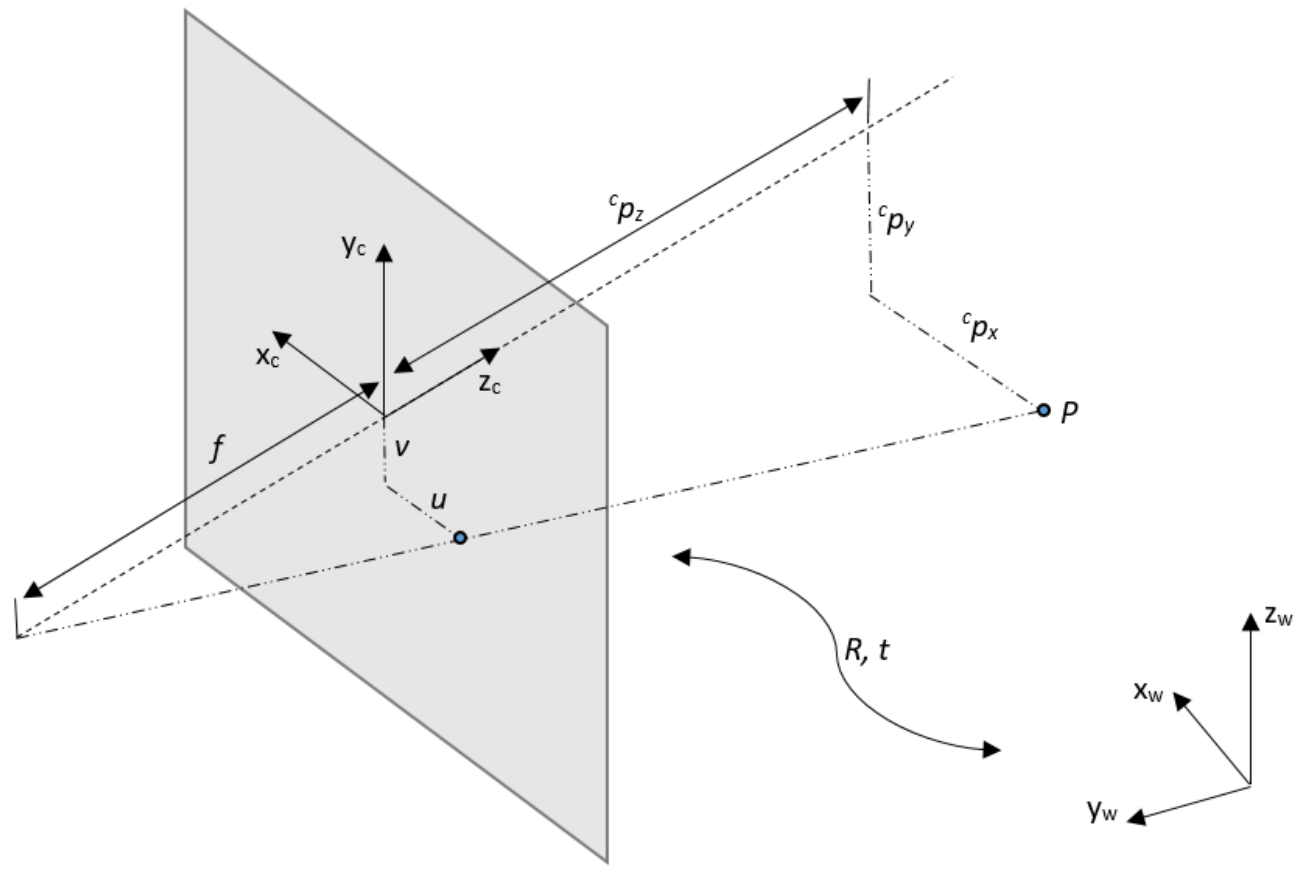

Fig. 1. Pin-hole camera model parameters.

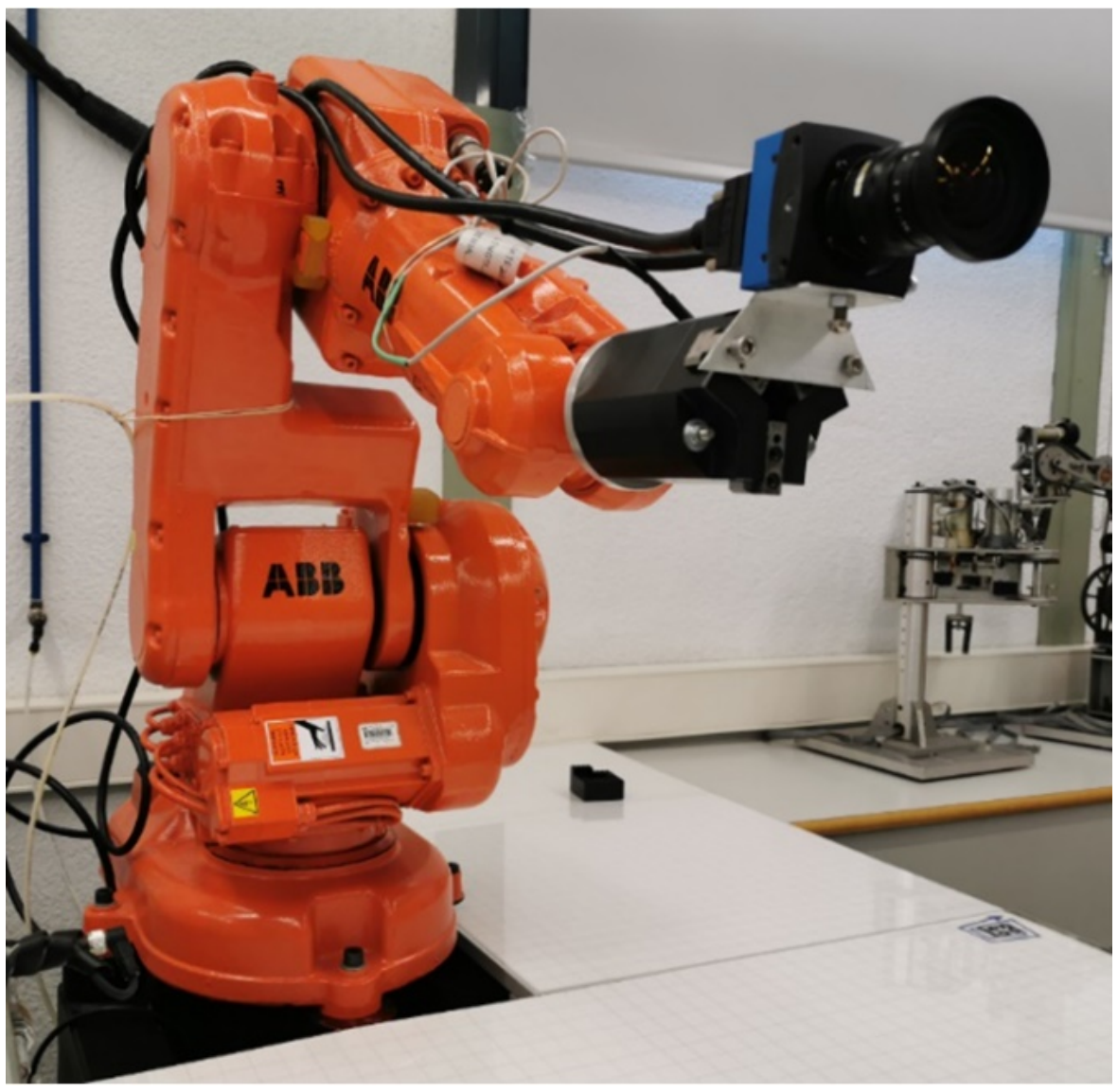

Fig. 2. Robot arm ABB IRB 140 with an EoSens® 12CXP+ camera of 4,096 × 3,072 pixels and sensor size of 23.04 × 23.04 mm active area with 18 mm lens with manual focus.

calibration process with a set of rules that help to improve the method. Section 2 analyses the camera calibration process using a two-dimensional template proposed by Zhang [4]. This analysis demonstrates that the interdependence of camera parameters means that the computed model cannot accurately represent the real camera in some cases. This inaccuracy arises in some conditions when if the final nonlinear minimization process computes all camera parameters at the same time. To demonstrate the inaccuracy of the camera calibration process, the camera is assembled on a robot arm in order to be able to accurately measure the extrinsic camera parameters. Computed camera parameters with a reprojection error close to zero are compared to the real ones so as to note the any discrepancy between them. Section 3 proposes a set of tests to calibrate cameras with guaranties of accurate results. First, intrinsic camera parameters are computed depending on the distance of camera to the calibration template and second, extrinsic parameters are computed alone to avoid the coupling between intrinsic and extrinsic camera parameters. Section 4 shows results that demonstrate the efficiency of the proposed method. Upon testing the proposed method, more accurate results are obtained because intrinsic camera parameters are isolated and they are constant in the iterative nonlinear minimization process. Paper ends with conclusions.

## 2 Analysis of the Existing Camera Calibration Process Using a 2D Template

Many current methods compute both camera parameters and lens distortion models. The most widely implemented is the method proposed by Zhang in his 2000 paper on camera calibration [4]. This method uses several images of a 2D chessboard template to compute the camera pin-hole model represented as:

$$s \cdot {}^{c}p = A \cdot [R \quad t] \cdot {}^{w}p \qquad (1)$$

where ${}^{c}\boldsymbol{p} = [p_u, p_v, 1]^T$ is the 2D coordinates in the camera frame of the 3D point in the scene represented as ${}^{w}\boldsymbol{p} = [p_x, p_y, p_z, 1]^T$ . As the camera model works with projective geometry, $s$ is an arbitrary scale factor. Rotation matrix $\boldsymbol{R}$ and translation vector $\boldsymbol{t}$ are extrinsic parameters that relate to the camera and world frames. Extrinsic camera parameters represent the location of the camera in the world. Matrix $\boldsymbol{A}$ contains intrinsic parameters as follows:

$$A = \begin{bmatrix} \alpha & \gamma & u_0 \\ 0 & \beta & v_0 \\ 0 & 0 & 1 \end{bmatrix} \qquad (2)$$

$\alpha$ and $\beta$ are the scale factors in the $u$ and $v$ camera axis. Both scale factors are defined with the focal length $f$ of the camera and the size of the camera sensor along the $u$ and $v$ axis. $\gamma$ is the skewness of axis $u$ and $v$ in the case they are not orthogonal. $u_0$ and $v_0$ are the coordinates of the principal point in the image plane. Figure 1 shows a diagram of the pin-hole camera model parameters.

The calibration process proposed by Zhang consists of several steps. First, $n$ images of a flat chessboard template are captured from different locations. At each location where the images were captured, the coordinates of the template points and their projection in the image are used to compute homographies between the template plane and camera plane. Second, combining elements of all homographies, intrinsic camera parameters are computed. Third, extrinsic camera parameters are computed for every location in which an image was taken. The final step of calibration consists of an iterative nonlinear minimization process to improve the computed intrinsic and extrinsic parameters, minimizing the reprojection error given by:

$$\sum_{i=1}^{n}\sum_{j=1}^{m}\left\| {}^{c}p_{ij} - p^\wedge\left(A, R_i, t_i, {}^{w}p_j\right)\right\|^2 \qquad (3)$$

where $n$ is the number of images, $m$ is the number of points in the chessboard template and $\boldsymbol{p}^\wedge(\boldsymbol{A},\boldsymbol{R}_i,\boldsymbol{t}_i,{}^{w}\boldsymbol{p}_j)$ is the projection with the estimated camera parameters $\boldsymbol{A}$, $\boldsymbol{R}_i$, $\boldsymbol{t}_i$, of calibration template point ${}^{w}\boldsymbol{p}_j$ in the image $i$ according with equation (1). The minimization of the reprojection error shown by (3) is an iterative nonlinear minimization prob-lem that is solved using the Levenberg-Marquardt algorithm. The aim is to reduce the distance that exists between detected points in images and projected points in images using the computed model.

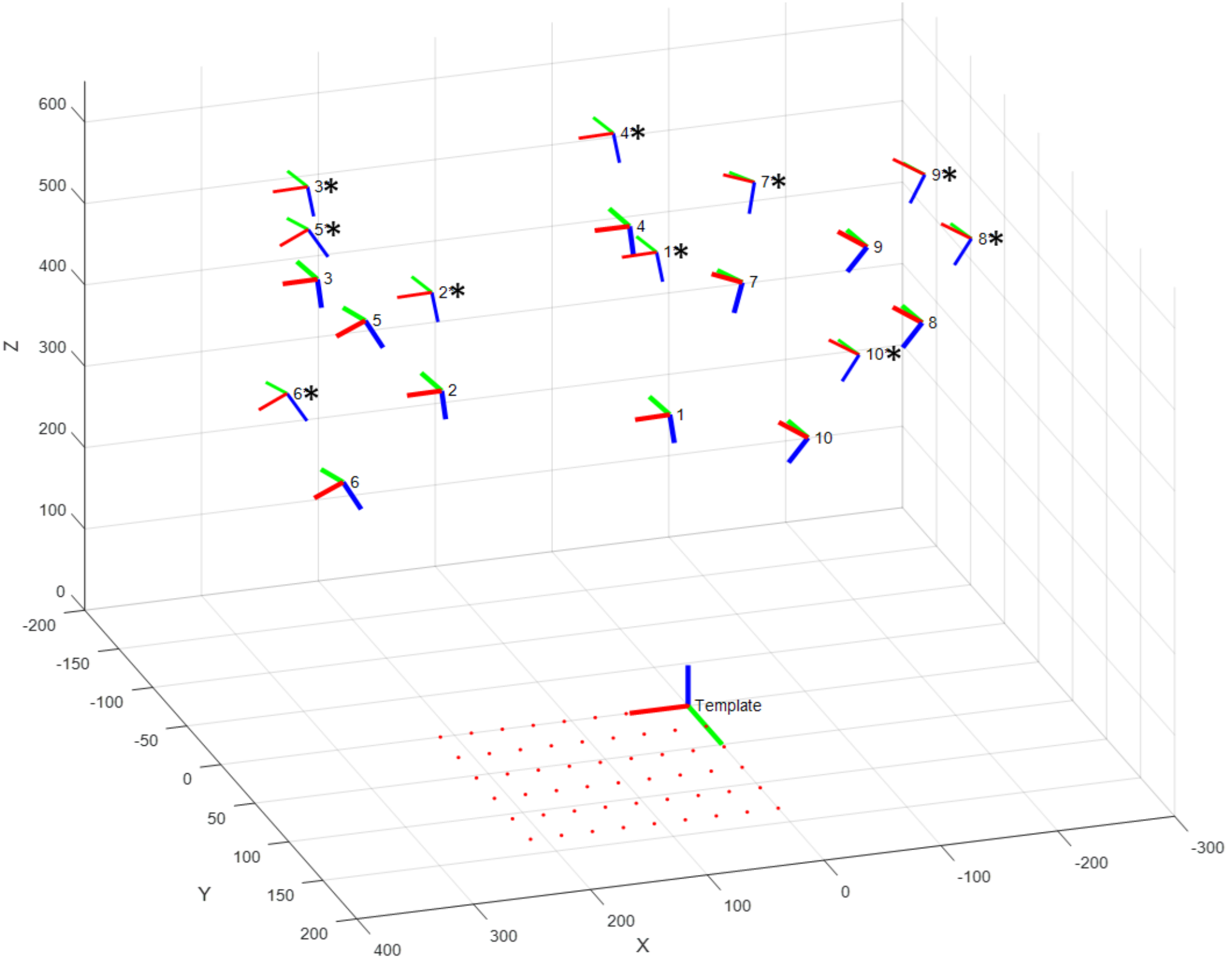

Fig. 3. Camera locations for 10 images. True camera locations have an asterisk. They are known because the camera is assembled on the robot arm. Computed camera locations are denoted with the number only. Computed camera locations are all closer to the template by about 100 mm.

## 2.1 Empirical test of the calibration process

Normally, intrinsic and extrinsic camera parameters are unknown and the result of the calibration process is considered valid because the reprojection error is close to zero. To verify the accuracy of computed camera parameters, an empirical experiment is performed. Figure 2 shows the setup of this experiment. It uses an EoSens® 12CXP+ camera of 4,096×3,072 pixels, sensor size of 23.04×23.04 mm active area and an 18mm manual focus lens. This camera is assembled on a robot arm ABB IRB 140 for calibration purposes. As the camera is on a robot arm, extrinsic camera parameters can be accurately known when an image is taken to calibrate it. This allows a comparison of the calibration algorithm results with the true values given by the location of the robot arm. Calibrating the camera with the widely accepted method, proposed by Zhang, which uses a 2D template, gives a valid solution because the reprojection error is close to zero. However, when computed extrinsic camera parameters are compared with the true real ones, they are consistently revealed to be incorrect.

The error in computed parameters is depicted in figures 3 and 4. The reprojection error defined in equation (3) has a mean value of 1.2092×e-09 with a standard deviation of 2.3201 pixels. The computed model projects all the template points in the images using equation (1) and 98% of distances between the projected and detected points in images is in a range of ±4.64 pixels. Since the image has a resolution of 4,096×3,072, 4.64 pixels represent an error of 0.1%, which is small enough to validate the computed model. In this case, since the camera location is known for each image, computed extrinsic parameters are compared with the true ones to validate the calibration result. Figure 3 shows the calibration stage, including the location of cameras and template points, both in the capturing stage and after the calibration process. Locations with an asterisk denote the true locations when images were captured with the camera on the robot arm. Locations without an asterisk show the location of the camera given by the calibration process. All computed camera locations are 100mm closer to the calibration template than the true location. Figure 4 shows the resulting calibration errors in location for coordinates x, y and z. Errors are computed using the true camera location as the reference frame. Errors in the x and y-axis are positive and negative in a range from -75mm to 75 mm. However, errors in Z coordinates show that the computed camera location is always closer to the calibration template than the true real location by a mean value of 100mm, within a range from 60 mm to 160 mm. This means that, in all cases, the computed camera location is closer to the calibration template than the true location. Consequently, values for scale factors α and β are biased.

To summarize, the calibrated model is considered valid because the reprojection error is close to cero, but empirical experiments show that there are consistent bias in camera location. Bias in camera location means bias in intrinsic camera parameters if reprojection error is close to cero. If the camera is calibrated using the currently well-established method, it is possible to arrive at a seemingly valid solution, but it does not represent the real camera.

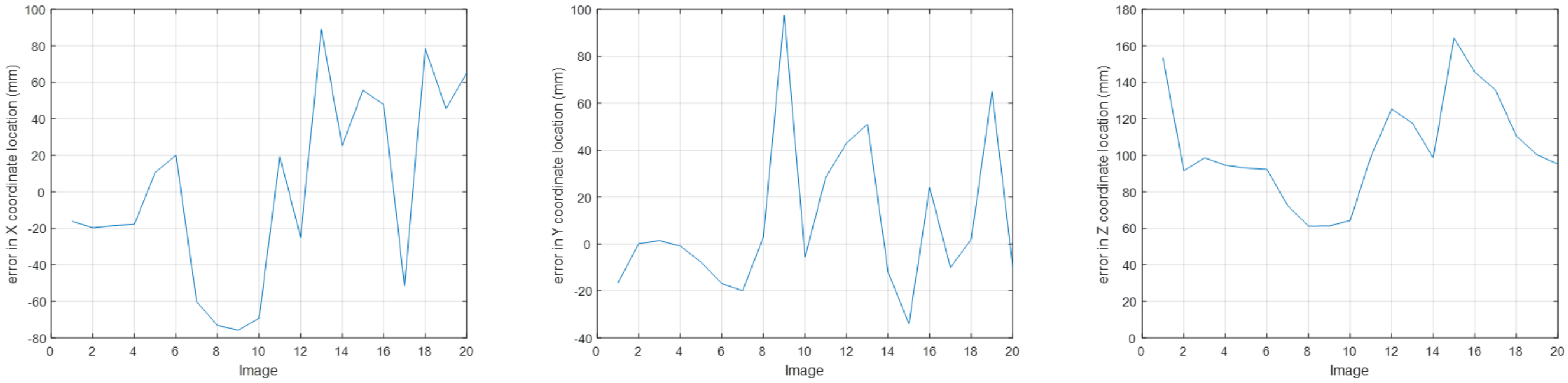

Fig. 4. Bias in computed camera locations. The reference frame is the true camera location obtained with the robot arm. Bias in X and Y-axis are positives and negatives, ranging from -75mm to 75 mm. However, bias in Z coordinates show that the computed camera location is always closer to the calibration template than the real location in 100mm as mean value within a range from 60 mm to 160 mm.

## 2.2 Why are the camera parameters biased?

Calibrating a pin-hole model of a camera with focus lenses is liable to failure. This is because the pin-hole model does not represent the focus camera behaviour accurately. A focus camera system adjusts lenses to focus on the chessboard template with every image. In consequence, focal length $f$ (expressed with the scale factors $\alpha$ and $\beta$ in the camera model) is not constant and a systematic error arises if the pin-hole model with a constant focal length is adjusted for different camera-chessboard distances.

To perform a precise camera calibration process, the computed model is only valid for a range of distances from the camera to the calibration object. Given a fixed focal length, the range of distances in front of and behind the focal plane within which objects appear sharp is called the depth of field. Assuming that applications need sharp images for a given circle of confusion, objects appear acceptably sharp in images when they are in the depth of field zone. If an object is out of focus, the focal length should change to obtain a sharp image of the object. In consequence, for a specific application, it is necessary to know the required range of distances between the camera and the object to perform a precise calibration. Out of this zone, the sharpness of objects will be poor and focal length will have to be changed to improve it.

The depth of field is controlled using the camera diaphragm aperture and the focal length. Given a constant diaphragm aperture, the rule is that the closer the object is, the smaller the depth of field is; conversely, the further the object is, the greater the depth of field is until it is infinite. The hyperfocal distance is considered when the object is far away from the camera and the depth of field is infinite. Therefore, if the calibration is performed with the template located at different distances from the camera, it could obtain biased results, as the focal length varies in each image to obtain sharp objects. Figure 5 illustrates this effect.

## 2.3 Where is the hyperfocal distance?

In figure 6 an analysis of rays that form the image passing through the camera lens is performed. The ideal projective ray that represents the pin-hole model is the projection of point $\boldsymbol{p}$ in the scene passing through the pin-hole of the camera. It is denoted in red. The ray going through the edge of the lens is deviated according with the curvature of the lens and it is denoted in green. The ratio between the angle of the incoming ray $\varphi$ and the angle of outcoming ray $\omega$ is constant for a distance to the centre of the lens. As the distance of the object to the lens varies, so do the angles $\varphi$ and $\omega$, as is shown in figure 6 for two distances. The sharpest image arises when the intersection of the projective ray with the ray coming from the edge of the lens coincides with the sensor plane, as is shown in figure 6. This point is represented as point $\boldsymbol{q}$ in that figure.

An analysis of the variation of the point $\boldsymbol{q}$ with the distance of the object to the camera will give the hyperfocal distance. The variation of point $\boldsymbol{q}$ gives different values of the focal length $f$ depending on the distance of the object to the camera. Two cases are shown in figure 6. Figure 6(a) shows the point $\boldsymbol{q}$ when the object is far away from the camera. In figure 6(b) the object is close to the camera. Depending on the distance of the object to the camera the angle of the incoming ray $\varphi$ varies until a point that this variation is insignificant for a circle of confusion. This point is where the hyperfocal distance is defined.

From a mathematical point of view, considering a coordinate system $XY$ with the origin in the intersection of the optical axis with the lens plane, coordinates of point $\boldsymbol{p}$ in figure 6 are $(d, a)$ where $d$ is the distance of point $\boldsymbol{p}$ to the lens, and $a$ is its distance to the optical axis. The equation of the ideal projective ray that represents the projection of point $\boldsymbol{p}$ in the scene going through the pin-hole of the camera is:

$$y = -\frac{a}{d} \cdot x \tag{4}$$

The equation of the projective ray of point $\boldsymbol{p}$ that goes through the edge of the lens is:

$$y = \frac{D-a}{d} \cdot x + D \tag{5}$$

where $D$ represents the radius of the lens. The angle of the incoming ray to the lens $\varphi$ is defined as:

$$\varphi = \tan^{-1}\left(\frac{d}{D-a}\right) \tag{6}$$

The angle of the outcoming ray $\omega$ is proportional to the angle of the incoming ray $\varphi$ as $\omega = k \cdot \varphi$ where $k$ depends on the lens. The equation of the outcoming ray of the lens is:

$$y = -\tan\left(\frac{\pi}{2} - \varpi\right) \cdot x + D \tag{7}$$

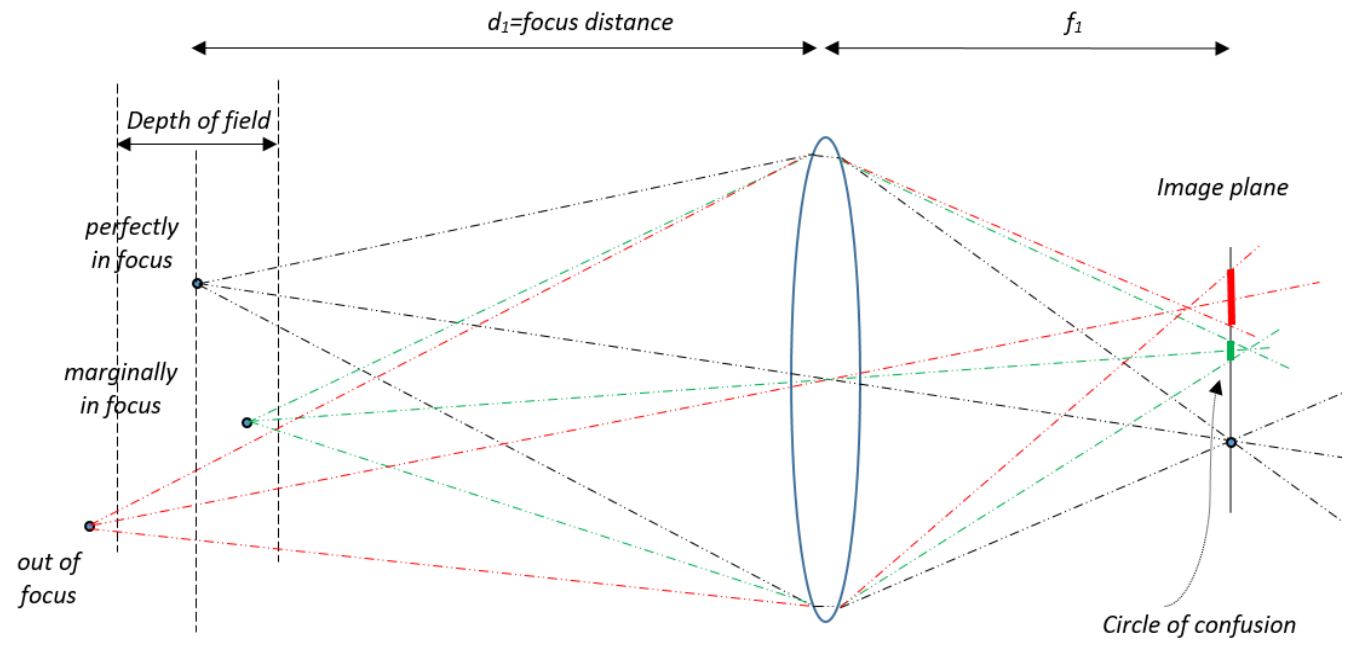

Fig. 5. Camera optics system adjusts lenses to focus the object in the scene. In consequence, focal length f and scale factors α and β are not constant and depend on the distance of the camera to the object. Assuming that applications need sharp images for a given circle of confusion, objects appear acceptably sharp in images when they are in the depth of field zone. Given a fixed focal length, the range of distances in front of and behind the focal plane is called depth of field, and objects in this zone appear sharp with no variation of focal length. If an object is out of focus, focal length should be changed so as to obtain a sharp image of the object.

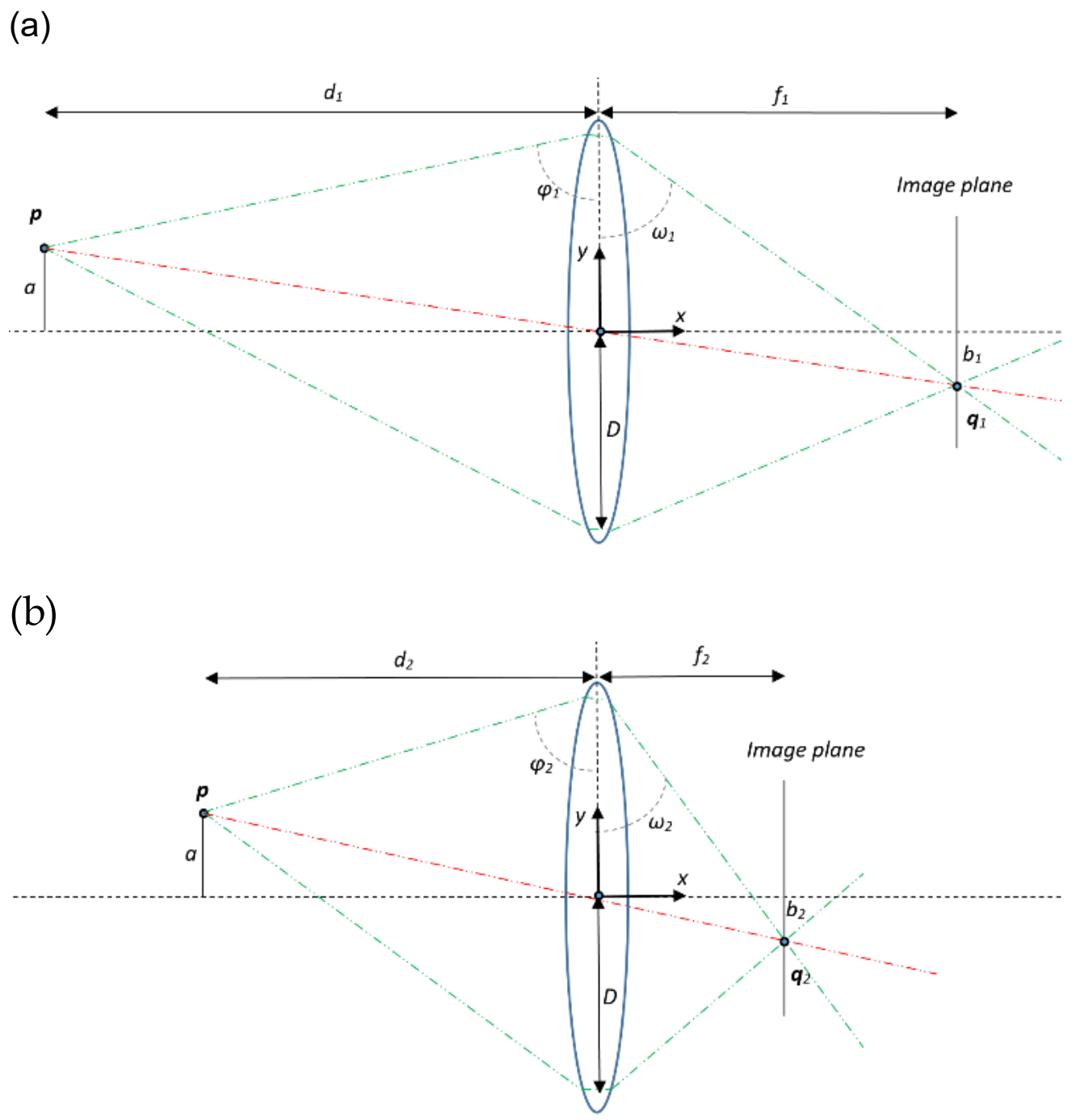

Fig. 6. The sharpest image appears when the intersection of the ideal pin-hole ray and the ray coming from the edge of the lens is in the camera sensor plane. The angle of the incoming ray φ and the outcoming ray ω varies with the distance of the object to the lens. Camera optics system adjusts lenses to focus the object in the scene.

The intersection of line defined in (4) and line defined in (7) gives the point $\boldsymbol{q}$ where the sensor plane must be positioned to obtain the sharpest image. Coordinates of point $\boldsymbol{q}$ are $(f, b)$, where $b$ represents the projection of the point $\boldsymbol{p}$ in the image plane and $f$ is the distance of the sensor to the lens defined as the focal length. Lines (4) and (7) intersection is given by:

$$-\frac{a}{d} \cdot x = -\tan\left(\frac{\pi}{2} - \varpi\right) \cdot x + D \tag{8}$$

The value of $x$ which solves (8) represents the focal length $f$. Therefore, the focal length that gives the sharpest image is computed as:

$$f = \frac{D}{\tan\left(\frac{\pi}{2} - \varpi\right) - \frac{a}{d}} \tag{9}$$

Expression (9) demonstrates that the focal length varies to obtain sharp images when the distance of the object to the camera changes. Moreover, the focal length depends on the sensor size. The size of the sensor is represented by the radius of the lens $D$ that illuminates it. The bigger the sensor is, the larger the value of $D$. Figure 7 represents how focal length varies with the distance of the object to the camera defined in equation (9). When the object is far away from the camera, variations of $f$ to obtain a sharp image are insignificant. Considering that the sharpness of the image is defined up to a circle of confusion, objects that are far away from the hyperfocal distance of the camera will be focused.

### 2.4 Why is the reprojection error close to zero and parameters are biased?

It's widely accepted that the result of the calibration process is valid if the reprojection error defined in equation (3), gives an error close to zero. However, an exhaustive analysis of the pin-hole camera model reveals that this assumption is not definitive.

A camera is a tightly coupled system in which errors of intrinsic parameters are compensated with incorrect values in the extrinsic parameters, and vice versa, to keep the reprojection error close to zero.

On the one hand, the camera focal length parameter $f$ is tightly coupled with the distance between the camera and the calibration template, as can be observed in figure 8. According to figure 8 the focal length and the distance of the object to the camera $p_z$, are related as:

$$\frac{v}{p_y} = \frac{f}{p_z} \tag{10}$$

Using coordinates $v$ and $p_y$ to compute camera parameters $f$ and $p_z$, the results are infinitely varied because the parameters are interdependent. According to equation (10) any pair $f$ and $p_z$, that satisfy the ratio $v/p_y$ is valid. In figure 8(a) and (b), given fixed $v$ and ${}^{c}p_x$, both different solutions of $f$ and ${}^{c}p_z$ are correct according to the criteria of the reprojection error, because both solutions satisfy it properly.

On the other hand, the camera location, the focal length and the principal point of the camera are also tightly coupled. Figure 9 shows this coupling. Both images show valid solutions for the calibration problem using the same coordinate points in the template $\boldsymbol{p}$ $(p_x, p_y)$ and the image $\boldsymbol{q}$ $(u, v)$ as the input data of the calibration process. In both cases, camera parameters and the calibrating data give a reprojection error equal to zero. Errors in the principal point are compensated by incorrect values in the camera location and vice versa. Moreover, biased values in the principal point and camera location have a direct effect on the camera focal length.

According to figure 9, given an optical axis perpendicular to the image plane, all camera parameters are represented by this optical axis. In this figure, two different focal lengths, which use the same calibration data, are considered valid because other camera parameters compensated this variation. Both solutions arise changing the optical axis, which is perpendicular to the image plane. There is a plane that represents all candidates to camera optical axis given a pair of points $\boldsymbol{p}$ and $\boldsymbol{q}$ to calibrate the camera. This plane is called the Sophia plane and it is represented in figure 9(b). This plane contains the ray that goes through the point $\boldsymbol{p}(p_x, p_y)$ to the point $\boldsymbol{q}(u, v)$ and all optical axis that represent different camera parameters. Depending on the chosen optical axis, camera parameters vary.

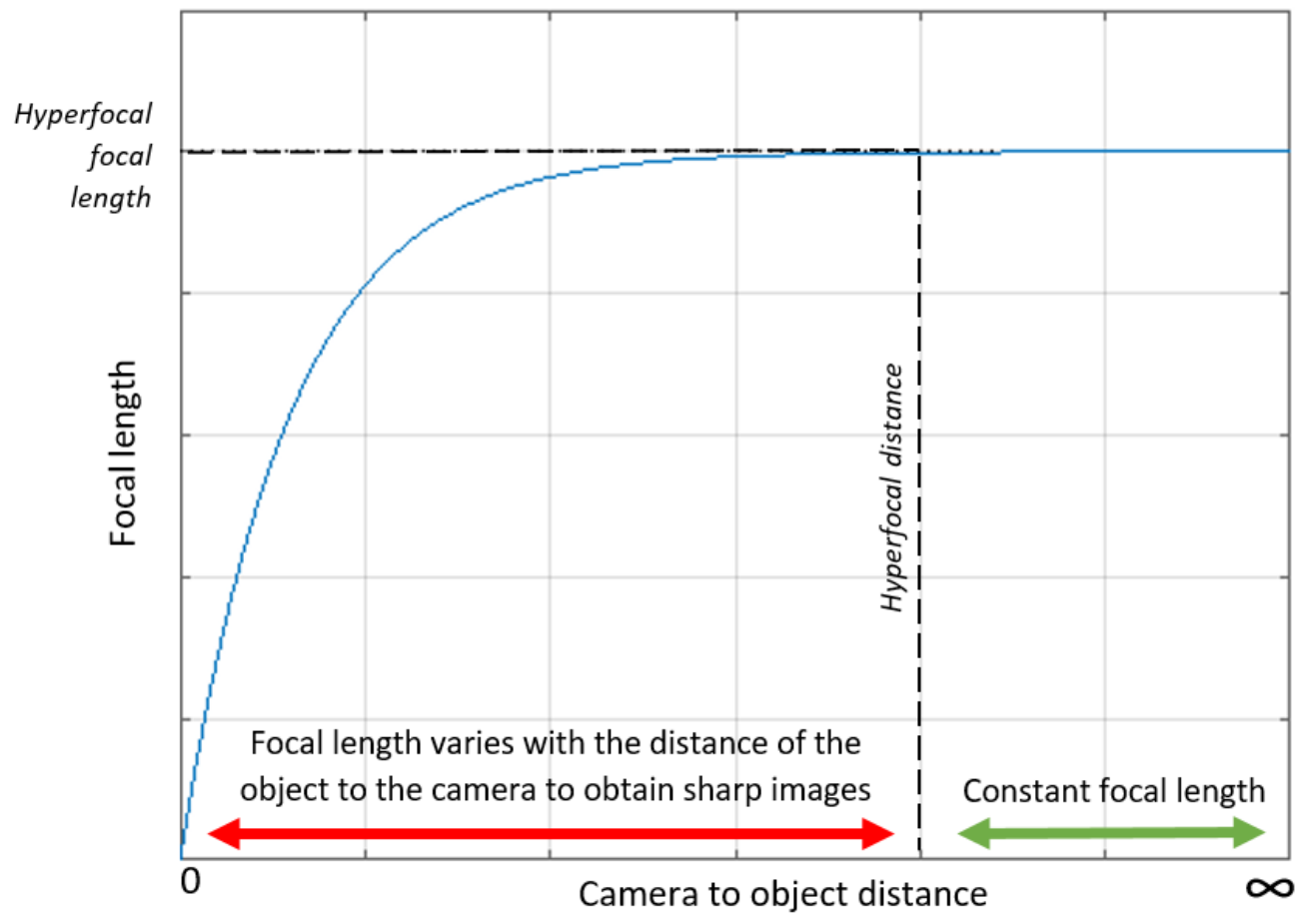

Fig. 7. Variation of the focal length with the distance of the object to the camera according with expression (9).

Camera parameters coupling has been demonstrated using just one pair of points $\boldsymbol{p}$ and $\boldsymbol{q}$ in the scene and in the image. When several points are used to calibrate the camera, several Sophia planes arise and the camera parameters are defined with the best optical axis that satisfy all Sophia planes. It is considered, that using several images of a calibration template with several calibration points, the true solution of camera parameters will be obtain. However, this assumption fails if the empirical experiment described in subsection 2.1 is performed. Results showed in figures 3 and 4 are biased. This empirical experiment demonstrates that using several calibration points in several images do not guarantee unbiased camera parameters. Moreover, this bias is not detected if camera parameters are validated using the reprojection error tool.

## 3 Proposed Camera Calibration Process using a 2D Template

Existing camera calibration methods perform two steps. First, an approximation of camera parameters is computed based on an algebraic solution. Second, an iterative nonlinear minimization problem improves the algebraic solution according to the criterion of the reprojection error, with the aim of computing the correct parameter values. Intrinsic and extrinsic parameters are updated in every iteration. It is assumed that using several images taken from different locations will condition the nonlinear minimization algorithm to achieve the right solution [4, 10].

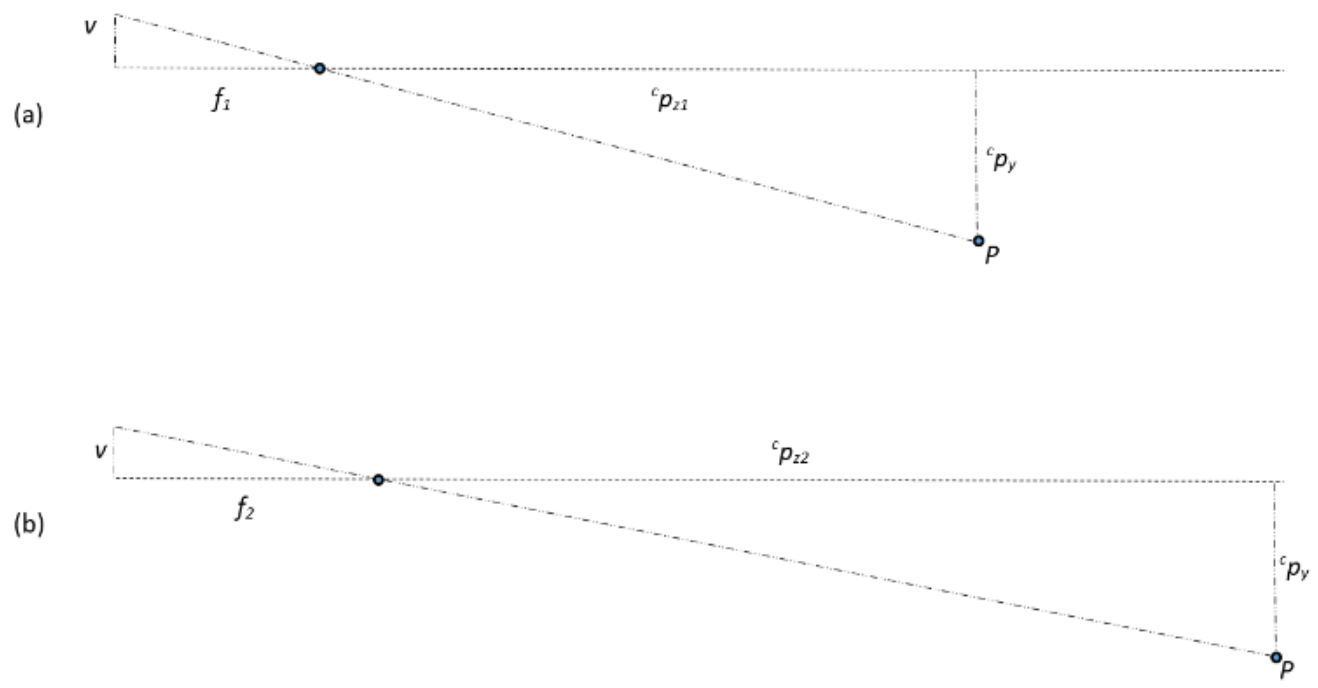

Fig. 8. Coupling between the camera focal length parameter *f* and the distance of the camera to the calibration template. Using coordinates *v* and $^cp_y$ to compute camera parameters *f* and $^cp_z$ has infinite solutions because both parameters are tightly coupled. In this figure, starting with the same value of *v* and $^cp_y$ both *f* and $^cp_z$ solutions (a) and (b) are valid.

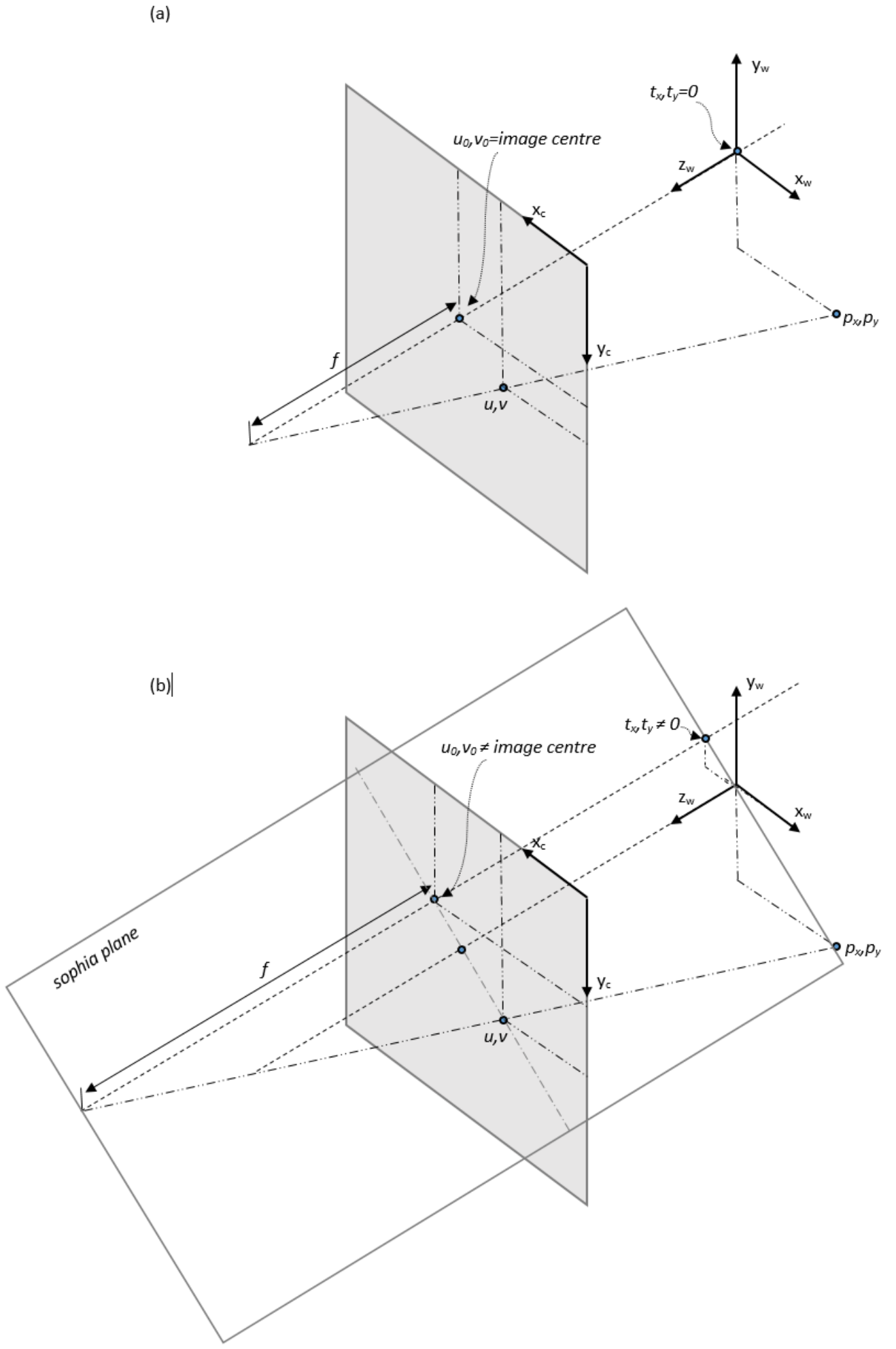

Fig. 9. Coupling between the principal point ($u_0$,$v_0$), focal length *f* and the extrinsic camera parameters $t_x$ and $t_y$. Changes in intrinsic camera parameters ($u_0$,$v_0$) are compensated with changes in the location of the camera $t_x$ and $t_y$ and focal length *f*. Figure (a) shows a solution for camera parameters and figure (b) shows another solution for camera parameters using the same data to compute camera parameters (coordinates points in the template ($p_x$, $p_y$) and coordinate points in the image ($u$, $v$)) Any optical axis that belongs to the Sophia plane gives a valid solution for the camera parameters. This is shown in figure (b). Using reprojection error as the tool to determine true camera parameters, all optical axis that belongs to the Sophia plane will be a valid solution.

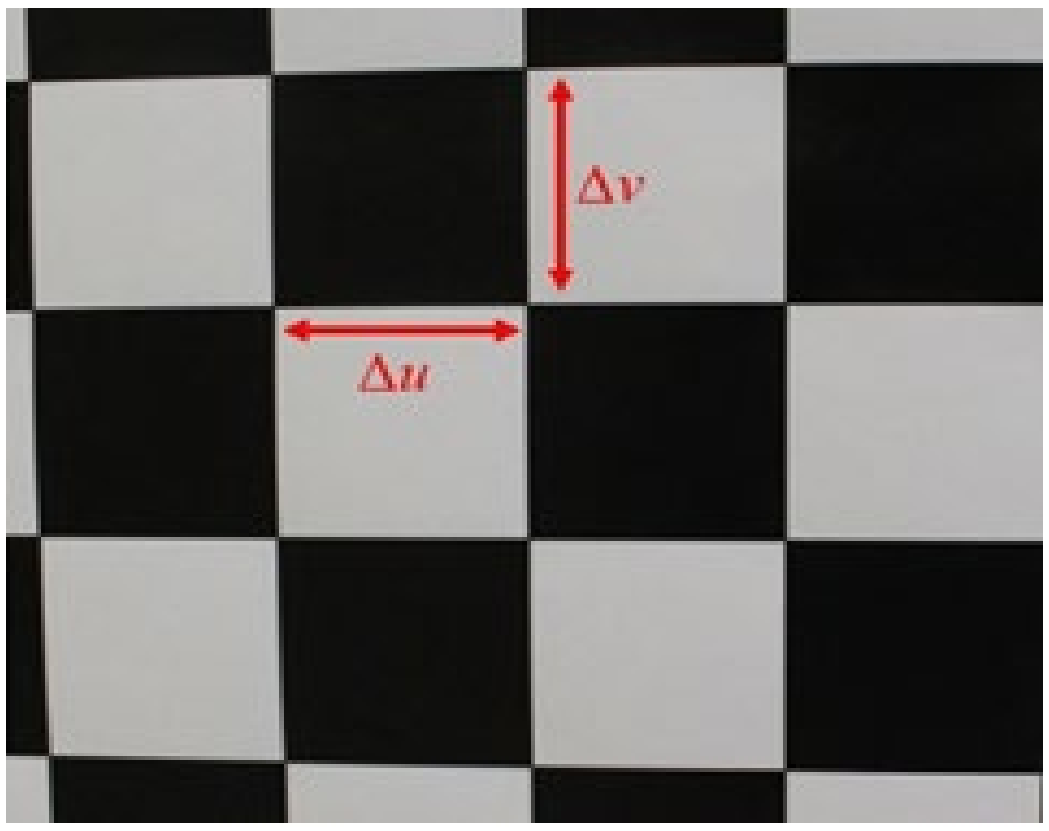

Fig. 10. Example of the image for the proposed method for calibrating the scale factors *α* and *β*. Resulting image and values *Δu* and *Δv*.

However, when a camera is calibrated using images from several locations, this assumption fails, as it was shown in the empirical experiment described in the previous section. As was shown in subsection 2.2.1, different distances from the camera to the calibration object produce changes in the focal length parameter to obtain sharp images. These variations in the focal length depend on the camera sensor size also, as was demonstrated in equation (9). Using the method proposed by Zhang, since the camera calibration process computes only one camera model with a fixed focal length, the interdependence between the camera parameters leads the nonlinear minimization process to get zero reprojection error that camouflages a biased result.

To improve the solution of the camera calibration process, it is necessary to define previously the range of distances from the camera to the calibration object where the focal length parameter will remain constant when getting a sharp image. This analysis will give the limits of the distances where the focal length varies in each image and the distance where the focal length keeps constant (i. e., where the hyperfocal distance starts).

Figure 7 represents these limits. The hyperfocal distance is the starting point from where focal length is constant although the distance between the object and the camera varies. If the distance between the object and the camera is smaller than the hyperfocal distance, the focal length is adjusted in each image to obtain sharp images.

Using the method proposed by Zhang in [4], if the calibration is done with images where the focal length varies (below the hyperfocal distance), all calibration images should be taken with similar distance between the calibration template and the camera in order to guarantee that the computed parameters are correct. If images taken at several distances in the range where the focal length varies are used, computed result using the method proposed by Zhang in [4] will be biased although the reprojection error is zero. In this case, since images are taken at several distances in the range where the focal length varies, several focal lengths should be computed depending on the distance of the camera to the calibration template. From the point of view of the application that uses the calibration results, calibration will be valid only for images that are taken in the same range of distances defined in the calibration process.

The aim of the proposed method is to identify this hyperfocal distance to ensure that the calibration process is correct. In the case where the calibration images are taken in a distance shorter than the hyperfocal distance, a method to compute several focal lengths that depends on the distance of the camera to the calibration template is proposed.

### 3.1 Identifying the hyperfocal distance and computing the scale factors *α* and *β*

The hyperfocal distance starts when the scale factors $a$ and $\beta$ are constant. As it was described in equation (2) scale factors $a$ and $\beta$ are defined with the focal length $f$ of the camera and the size of the camera sensor along the $u$ and $v$ axis. Using the projective geometry property described in figure 8 the coordinate $v$ of a point in the image is:

$$v = \frac{p_y}{p_z} \cdot \beta \tag{11}$$

where $\beta$ corresponds to the scale factor in the $v$ axis denoted with $f$ in figure 8. Two points in a template plane in the scene that is parallel to the image plane have the same $p_z$ and $p_{y1}$, $p_{y2}$, respectively. These two points will project the image in coordinates $v_1$ and $v_2$. The increment between the two coordinates in the $v$ axis is:

$$v_1 - v_2 = \frac{p_{y1}}{p_z} \cdot \beta - \frac{p_{y2}}{p_z} \cdot \beta = \left(p_{y1} - p_{y2}\right) \frac{\beta}{p_z} \tag{12}$$

$p_z$ is equal at both points because the template scene plane is parallel to the image plane. Several images of the same two points in the calibration template at different distances will result in different increments in the image and different scale factors.

$$\Delta v_i = \frac{\Delta t \cdot \beta_i}{d_i} \tag{13}$$

where the camera distance $p_z$ is denoted with $d_i$, and $\Delta v_i=(v_1-v_2)$, $i=1...n$ where $n$ is the number of images. $\beta_i$ corresponds to the scale factor in each distance of the camera to the calibration template. $\Delta t$ is the constant increment between adjacent points in a scene. Arranging the data in a matrix form, the following expression arises:

$$\begin{bmatrix} \beta_1 \\ \beta_2 \\ \dots \\ \beta_n \end{bmatrix} = \frac{1}{\Delta t} \cdot \begin{bmatrix} \Delta v_1 \cdot d_1 \\ \Delta v_2 \cdot d_2 \\ \dots \\ \Delta v_n \cdot d_n \end{bmatrix} \tag{14}$$

In the case that $a$ corresponds to the $u$ axis, the deduction is similar and the equation (14) becomes:

$$\begin{bmatrix} \alpha_1 \\ \alpha_2 \\ \dots \\ \alpha_n \end{bmatrix} = \frac{1}{\Delta t} \cdot \begin{bmatrix} \Delta u_1 \cdot d_1 \\ \Delta u_2 \cdot d_2 \\ \dots \\ \Delta u_n \cdot d_n \end{bmatrix} \tag{15}$$

If elements of vector $a_i$ or $\beta_i$ are plotted, a figure similar to the figure 7 appears that is very useful to decide where is the hyperfocal distance.

To compute the scale factors $\alpha_i$ or $\beta_i$ of the camera axis $u$ and $v$ using expressions (14) and (15), we propose a simple empirical experiment. Several images of a planar chessboard template are taken, in which the image plane is as parallel as possible to the chessboard plane. It is important to know the distance from the camera to the chessboard plane when the image is captured.

Figure 10 shows an example of the resulting image. Here, $\Delta u$ and $\Delta v$ are defined. The increment between adjacent points of the chessboard calibration template is $\Delta t$ and $d_i$ is the distance from the camera to the chessboard. The corners of the black and white boxes in the images are detected and the increments between the adjacent points are computed. The increments along the horizontal axis $u$ are denoted with $\Delta u_{i,j}$ and the ones in the vertical axis $v$ are denoted with $\Delta v_{i,j}$., where $i,j$ represent the increment $j$ of adjacent points in the image $i$. If the image has no distortion, the values of $\Delta u_{i,j}$ will be similar and their mean values will be represented with $\Delta u_i$. Likewise, in the $v$ axis, $\Delta v_i$ denotes the value of the increment along the $v$ axis in image $i$. Further, if the pixels are square, $\Delta u_i$ will be equal to $\Delta v_i$ and the scale factors $\alpha_i$ or $\beta_i$ will be equal.

In the case of distorted images, $\Delta u_i$ and $\Delta v_i$ is computed using the detected points in the centre of the image only, instead of using points all over the image. In images with radial distortion, points close to the centre of the image have an insignificant deviation from the undistorted position. If all points in the image are used, distortion could be corrected before computing the increments of $\Delta u_i$ and $\Delta v_i$ using the methods proposed by Ricolfe-Viala, [20, 21] or Zhu [22]. However, improvement on the results when distortion is corrected is negligible and therefore unnecessary.

## 3.2 Computing camera model with several scale factors

Once the experiment described in previous subsection is performed, it is possible to define the hyperphocal distance that establishes the limit where the focal length varies or it is constant. If calibration is done with images taken at distances where the scale factors are constant, the method proposed by Zhang in [4] could work properly. In case that the calibration process is done with images taken at distances where the scale factors vary, it is necessary to compute extrinsic parameters considering that the scale factors are different in each image.

This section describes a variation of the method proposed by Zhang in [4] to compute the extrinsic camera parameters using images taken at distances where the scale factors vary. The process consists of a nonlinear minimization process to adjust camera parameters to a set of data but in this case, computing only the principal point ($u_0$, $v_0$), the skewness and the extrinsic camera parameters and considering the scale factors as constant valued computed with the method described in previous subsection.

To perform a nonlinear minimization process, an initial guess is necessary. The scale factors $\alpha_i$ and $\beta_i$ for several distances $d_i$ are computed with the specific trial described in previous subsection. The image principal point parameters ($u_0$, $v_0$) is initialized to the centre of the image. This initialization does not disturb the final result because the nonlinear minimization process will obtain the right solution taking into account that the focal length is constant. It will find out the correct optical axis that satisfy the focal length in each image according with the figure 9. As it was said before, a set of camera parameters are defined with and optical axis and in this case, the constant value of the focal length will force the nonlinear minimization to compute the best optical axis for the given focal length. The skewness $\gamma$ is initialized to zero and the extrinsic camera parameters are obtained as follows.

Using initial values of intrinsic camera parameters, the matrix $\boldsymbol{A}$ defined in (2) has a specific value for each camera distance $d_i$. Moreover, coordinates of calibration template points have $z=0$ and the relation between the camera plane and the calibration template plane is defined with an homography $\boldsymbol{H}$. Equation (1) is rewritten as:

$$^{c}p = A_i\cdot[R_i \quad t_i]\cdot{}^{w}p \tag{16}$$

$$\begin{bmatrix} u \\ v \\ 1 \end{bmatrix} = A_i\cdot\begin{bmatrix} r_{11} & r_{12} & r_{13} & t_x \\ r_{21} & r_{22} & r_{23} & t_y \\ r_{31} & r_{32} & r_{33} & t_z \end{bmatrix}\cdot\begin{bmatrix} x \\ y \\ 0 \\ 1 \end{bmatrix} = A_i\cdot[r_1 \quad r_2 \quad t]\cdot\begin{bmatrix} x \\ y \\ 1 \end{bmatrix} \tag{17}$$

$$^{c}p = [h_1 \quad h_2 \quad h_3]\cdot{}^{w}p = H_i\cdot{}^{w}p \tag{18}$$

To compute the homography $\boldsymbol{H_i}$, a technique based on nonlinear least squares proposed in [4] is used. Let $\boldsymbol{c}=[\boldsymbol{h_1}\ \boldsymbol{h_2}\ \boldsymbol{h_3}]^T$, the equation (18) can be rewritten as:

$$\begin{bmatrix} {}^{w}p^T & 0^T & u\cdot{}^{w}p^T \\ 0^T & {}^{w}p^T & v\cdot{}^{w}p^T \end{bmatrix}\cdot c = 0 \tag{19}$$

With $n$ points, $n$ equation (19) arise which can be written in matrix equation as $\boldsymbol{L}\cdot\boldsymbol{c}=\boldsymbol{0}$ where $\boldsymbol{L}$ is a $2\cdot n\times 12$ matrix. The solution is the right singular vector of $\boldsymbol{L}$ associated with the smallest singular value, or the eigenvector, of $\boldsymbol{L}^T\boldsymbol{L}$ associated with the smallest eigenvalue. Matrix $\boldsymbol{L}$ is poorly conditioned because some elements are a constant 1, others are in pixels and others are in millimetres. Better results are obtained by performing a simple data normalization process, as proposed in Hartley [24].

Once the homography $\boldsymbol{H_i}$ is computed and using the intrinsic camera parameters arranged in a matrix $\boldsymbol{A_i}$, extrinsic camera parameters are computed as:

$$\begin{aligned} r_{1i} &= \rho\cdot A_i^{-1}\cdot h_{1i} \\ r_{2i} &= \rho\cdot A_i^{-1}\cdot h_{2i} \\ r_{3i} &= r_{1i}\times r_{2i} \\ t_i &= \rho\cdot A_i^{-1}\cdot h_{3i} \end{aligned} \tag{20}$$

Because of the noise, computed elements of $\boldsymbol{R}$ will not satisfy the properties of a rotation matrix. To approximate the computed 3×3 matrix $\boldsymbol{Q}$ to the best rotation matrix $\boldsymbol{R}$, the singular value decomposition of $\boldsymbol{Q}$ is necessary. Given the singular value decomposition $\boldsymbol{Q}=\boldsymbol{U}\cdot\boldsymbol{S}\cdot\boldsymbol{V}$, the best rotation matrix is $\boldsymbol{R}=\boldsymbol{U}\cdot\boldsymbol{V}^T$. For further reference to matrix computation, see Golub and Loan [25].

The maximum likelihood estimation is the nonlinear minimization process that improves computed results, re-ducing the geometric error between detected point coordinates in images $^{c}\boldsymbol{p}_{ij}$ and projected points with the computed camera model defined as $^{c}\boldsymbol{p}_{ij}\boldsymbol{\wedge}(\boldsymbol{A}_i, \boldsymbol{R}_i, \boldsymbol{t}_i, {}^{w}\boldsymbol{p}_j)$. The maximum likelihood estimation is computed by minimizing the following function:

$$\sum_{i=1}^{n}\sum_{j=1}^{m}\left\| {}^{c}p_{ij} - p\wedge\left(A_i, R_i, t_i, {}^{w}p_j\right)\right\|^2 \tag{21}$$

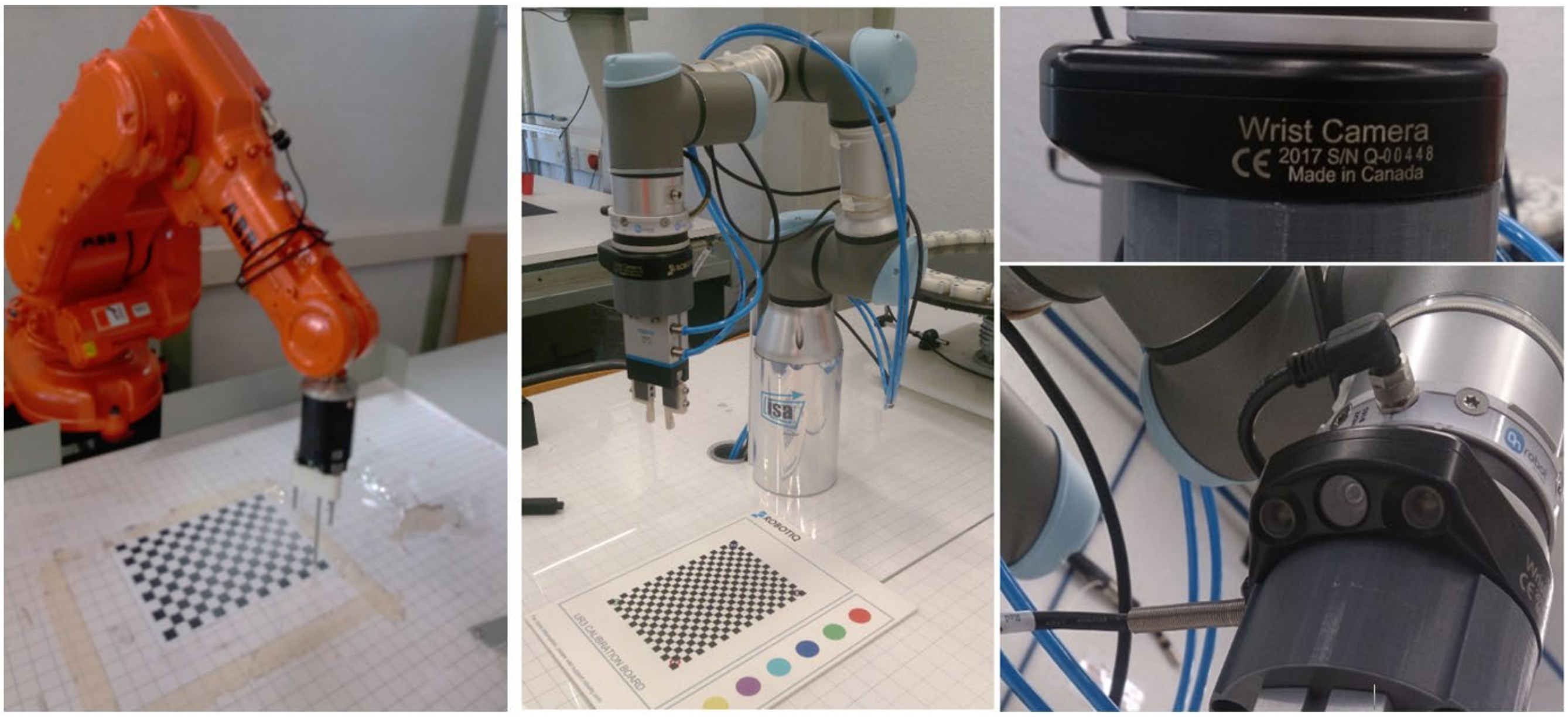

Fig. 11. Robot arm ABB IRB 140 with the measuring tool defining 3D coordinates of template control points. Robotiq wrist camera mounted in collaborative robot Universal Robot UR3.

Rotation matrix $\boldsymbol{R}_i$ is expressed with three parameters using the Rodrigues formula [26]. This nonlinear minimization problem is solved with the Levenberg-Marquardt algorithm using the values of $\boldsymbol{A}_i$, $\boldsymbol{R}_i$ and $\boldsymbol{t}_i$ computed previously as the searching starting point. Notice that several matrices $\boldsymbol{A}_i$ of intrinsic camera parameters exist that depend on the distance $d_i$. In this case, searching parameters are extrinsic camera parameters $\boldsymbol{R}_i$ and $\boldsymbol{t}_i$, the principal point ($u_0$, $v_0$) and the skewness $\gamma$. Scale factors $a_i$ and $\beta_i$ in matrices $\boldsymbol{A}_i$ remain constant to avoid the intrinsic and extrinsic camera parameters influencing each other which result in a biased solution at the end of the minimization process. Moreover, constant values of scale factors $a_i$ and $\beta_i$ force the searching process to find out the finest optical axis in the Sophia plane that best represents the camera model according to figure 9. When the minimization process is finished, reprojection error will be close to cero but in this case, it has been computed with a constant value of the focal length or scale factors $a_i$ and $\beta_i$.

### 3.3 Camera lens distortion

Most of the authors of papers about camera lens distortion report that distortion function is dominated by radial components if the image distortion is small [21, 22, 26]. A second order radial distortion model between the distorted point in the image $^{c}\boldsymbol{p}^{*}$ and the correct one $^{c}\boldsymbol{p}$ is defined as

$$^{c}p = {}^{c}p^{*} + \delta \tag{22}$$

such as

$$\begin{aligned}\delta_u &= \Delta u_d \cdot \left(k_1 \cdot r_u^2 + k_2 \cdot r_u^4\right)\\ \delta_v &= \Delta v_d \cdot \left(k_1 \cdot r_v^2 + k_2 \cdot r_v^4\right)\end{aligned} \tag{23}$$

where $r^2$ is the distance between the distorted point coordinate and the principal point. $\Delta u_d = u_d - u_0$, $\Delta v_d = v_d - u_0$. $r$ is computed as $r^2 = \Delta u_d^2 + \Delta v_d^2$.

Similar to the method proposed by Zhang in [4], the camera lens distortion model defined in (23) is included in the maximum likelihood estimation and equation (21) is extended taking into account radial distortion parameters:

$$\sum_{i=1}^{n}\sum_{j=1}^{m}\left\| {}^{c}p_{ij} - p\wedge\left(A_i, k_1, k_2, R_i, t_i, {}^{w}p_j\right)\right\|^2 \tag{24}$$

As before, scale factors $a_i$ and $\beta_i$ of matrices $\boldsymbol{A}_i$ remain constant to avoid the intrinsic and extrinsic camera parameters influencing each other. Maximum likelihood estimation computes extrinsic camera parameters $\boldsymbol{R}_i$ and $\boldsymbol{t}_i$, and the principal point ($u_0$, $v_0$), the skewness $\gamma$ and the distortion parameters $k_1$, $k_2$ as intrinsic camera parameters. $k_1$, $k_2$ are initialized to zero.

## 4 Experimental results

To demonstrate the influence of the focus in the camera calibration process, two set of images are captured using two cameras with different features assembled on two robot arms. The main difference between both cameras is the sensor size to verify that in bigger sensors, the hyperfocal distance is further away from the camera than in smaller ones as was defined in (9). Therefore, the risk of capturing images with different focal lengths for calibration purposes increases in cameras with bigger sensors. In consequence, if only one focal length is computed in the calibration process, biased parameters will be computed. In these cases, several camera models with different focal lengths are necessary to represent the focusing process of the camera when the distance of the camera to the calibration template varies under the hyperfocal distance.

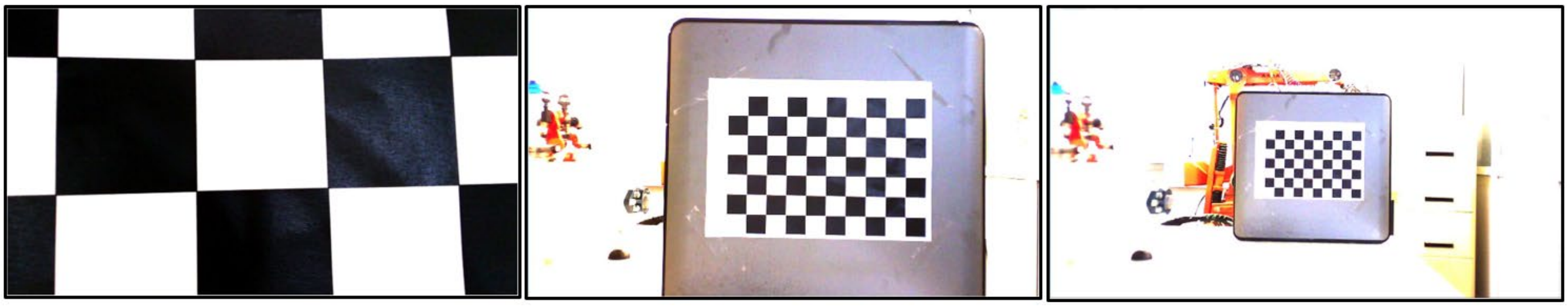

Fig. 12. Images to calibrate the scale factors and principal point with the proposed method with the Robotiq wrist camera, 100mm, 700mm, 1300mm, distance between camera and calibration template.

In our experiment, one camera is a Robotiq wrist camera, 1279×724 pixels with electrically actuated focus, assembled on a collaborative industrial robot UR3. The other one is an EoSens® 12CXP+ of 4,096 × 3,072 pixels, 23.04 × 23.04 mm active area with an 18 mm lens with manual focus, mounted on an ABB IRB 140. Both cameras are assembled on robot arms so as to establish the extrinsic camera parameters and to compare the computed values with the known true ones. It is assumed that better extrinsic parameters will result in more accurate intrinsic parameters with similar reprojection errors. The reprojection error is also computed to check if model is correct from a geometric point of view.

Camera location is obtained by using the location of the end of the robot arm that is provided by its control unit. With this location, it is possible to compute a transformation matrix that transforms the location of the robot arm into the location of the camera image frame or the measuring tool. Figure 11 shows an image of the robot arm ABB IRB 140 with the measuring tool, obtaining 3D coordinates of the calibration template points. The same operation is performed with the collaborative industrial robot UR3. Camera and template locations are in the robot base coordinate frame. No simulated data is used because it does not represent the real world. With computer simulations, data is generated with models that have a constant focal length with independence of the distance of the camera to the calibration template. Under the simulation umbrella, everything fits perfectly (even with Gaussian noise), because real camera behaviours such as focus features, are not considered in the theoretical model. In this case, only real data is used to demonstrate the influence of the camera focus in the calibration process.

### 4.1 Computing the scale factors $\alpha_i$ and $\beta_i$.

To compute the scale factors $\alpha_i$ and $\beta_i$ using the method proposed in subsection III.A., several images of the calibration chessboard template are captured, taking into account that the image plane should be as parallel as possible to the template plane. In addition, the distance of the camera to the chessboard template is known for each image. Figure 12 shows selected images that were captured to compute the scale factors $\alpha_i$ and $\beta_i$ of both cameras. Chessboard corners were detected using the Harris corner detection algorithm implemented in openCV. Increments $\Delta u_i$ and $\Delta v_i$ were computed as described in subsection III.A.

Scale factors $\alpha_i$ and $\beta_i$ for each camera-template distance are computed with equations (14) and (15). As it was assumed, the focal length is not constant in all distances as it is shown in figure 13. Black crosses represent the values of the elements of the vector $\alpha_i$ in (15). With a real camera, focal length is adjusted when the distance of the camera to the template changes to capture sharp objects. In consequence, scale factors $\alpha_i$ or $\beta_i$ do not have a constant value, except when the focus is set in the hyperfocal distance. When the focus is in the hyperfocal distance, the depth of field is at its maximum and the focal length does not change to capture sharp images. This effect is illustrated with the real values computed with the cameras. Figures 13(a) and 13(b) use crosses to show the elements of vector $\alpha_i$ for both the Robotiq wrist camera and the EoSens® 12CXP+ camera respectively. Analysing the results, three zones appear when the scale factor is plotted versus the distance of the camera to the template. Zone 1 is defined by a variation of the focal length with each image to obtain sharp images. Zone 2 is defined by the constant value of the focal length: the focus is set in the hyperfocal distance and without changing the focal length, images are sharp in the image although the distance of the camera to the object changes. Zone 3 is defined by variations in the scale factor because the template is far away from the camera and the poor quality of the calibration template in the image does not allow accurate detection of the image's control points.

With this experiment, it is possible to obtain the range of distances at which the camera will work properly with a constant focal length defined using Zone 2. In addition, if the camera is closer to the object and it is working in Zone 1, different values of the focal length will help to obtain accurate camera parameters. Moreover, Zone 3 will give the distance of the camera to objects in which the accuracy is reduced because the image resolution is not enough to detect object details with precision. Zone 3 will be conditioned by the size of the chessboard calibration template. The details of the objects than can be detected will be defined by the size of the squares in the chessboard template (in this experiment, one size measures 25 mm).

Zone 1 of the Robotiq wrist camera is between 0 and 150 mm. Zone 2, where the focal length is constant, is when objects are between 150 mm to 1200 mm away from the camera. If objects of 25 mm are more than 1200 mm away from the camera, the detection of their details is noisy, meaning that the measurements are not accurate. On the other side, with the EoSens® 12CXP+ camera, zone 1 is from 0 mm to 700 mm and zone 2 is from 700 mm to 2700mm, approximately.

Figures 13(a) and 13(b) show the computed values of the scale factor $\alpha_i$ using different methods. As mentioned before, black crosses represent the values of the elements of the vector $a_i$ in (15). The black line is the mean value of the elements of the vector $a_i$ in zone 2, which corresponds to the scale factor $a$ computed with no variation of focal length because the lens is focusing in the hypefocal distance. The red line shows the computed value using the method based on the 2D calibration template proposed by Zhang[4] when the maximum likelihood estimation (MLE) ends. The blue line shows the value computed with the algebraic solution from the Zhang method before it is improved with the maximum likelihood estimation. Calibration with Zhang method is performed with 15 images taken from several locations. Similar results are computed for the scale factor $\beta$. Values calculated using each method are shown in table 2.

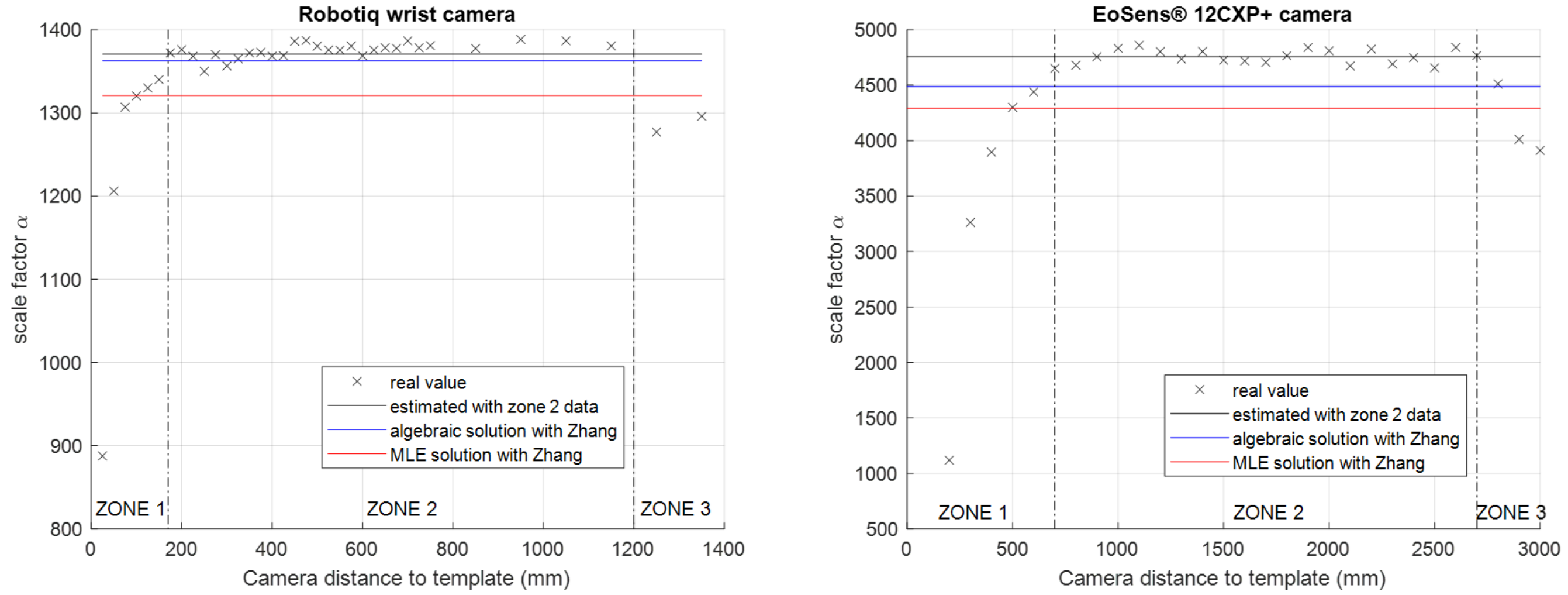

Fig. 13. (a). Results for the Robotiq wrist camera assembled on a collaborative industrial robot UR3. (b) Results for the EoSens® 12CXP+ camera assembled on an industrial robot arm ABB IRB 140. Black crosses represent the result of equation (15). Each black cross is the scale factor αi computed with the increments of each image $\Delta u_i$ and the distance di from which it was taken. The black line represents the mean value of all values in zone 2 (equations (14) and (15)). The red line shows the computed value using Zhang's method when the maximum likelihood estimation (MLE) ends. The blue line shows the computed value of the algebraic solution of the Zhang method before it is improved with the maximum likelihood estimation.

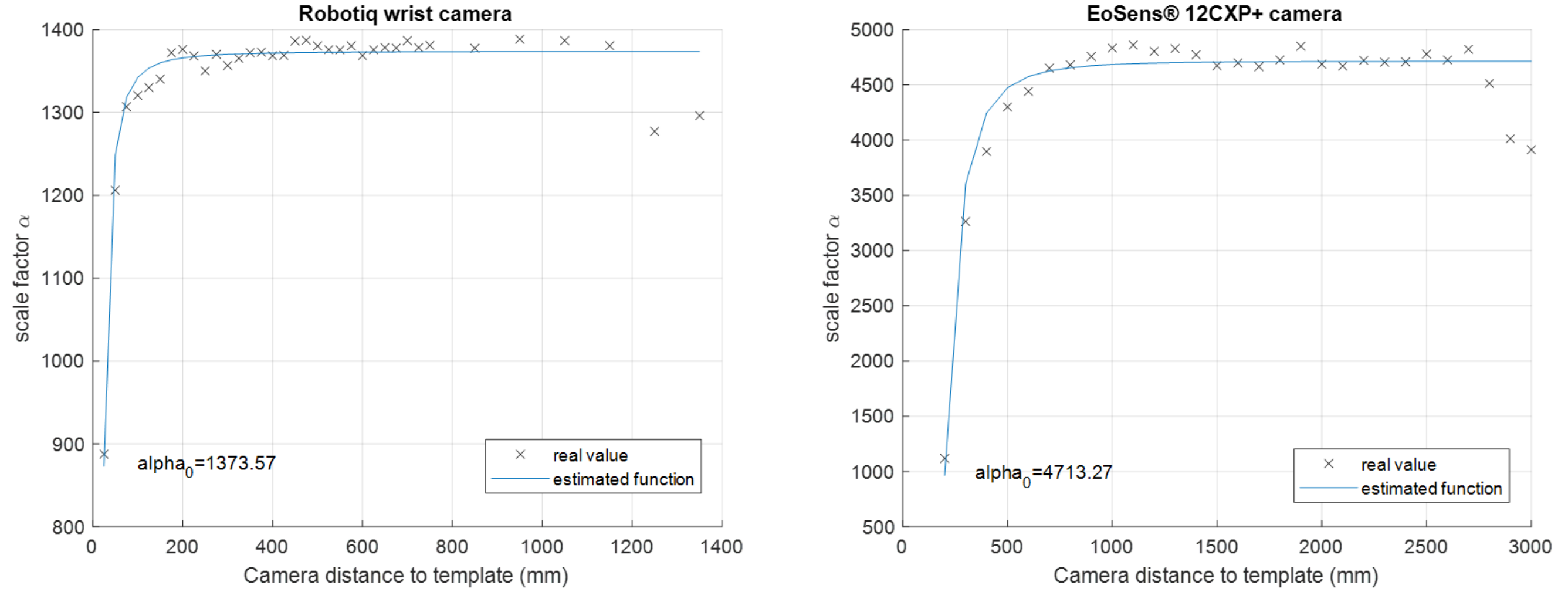

Fig. 14. Adjusting the camera scale factor to the function defined in (25) (a). Results for the Robotiq wrist camera. (b) Results for the EoSens® 12CXP+ camera.

If a continuous camera model were necessary to compute the focal length of the camera according with its distance to the object, a function can be empirically adjusted to data in figure 13 as follows:

$$\alpha = -\frac{k_f}{d^2} + \alpha_0 \qquad (25)$$

where $a$ is the scale factor for a distance camera-object defined with $d$ and $a_0$ is the scale factor when the focus is set to the hyperphocal distance. Parameters $k_f$ and $a_0$ can be adjusted using the least squares technique. Figure 14 shows the results of adjusting scale factor data of both cameras to the function defined in (26). Similar expression is computed for $\beta$.

## 4.2 Computing the complete camera model.

Using the scale factors $a_i$ or $\beta_i$ computed in previous subsection, it is possible to compute a complete set of camera

(a)

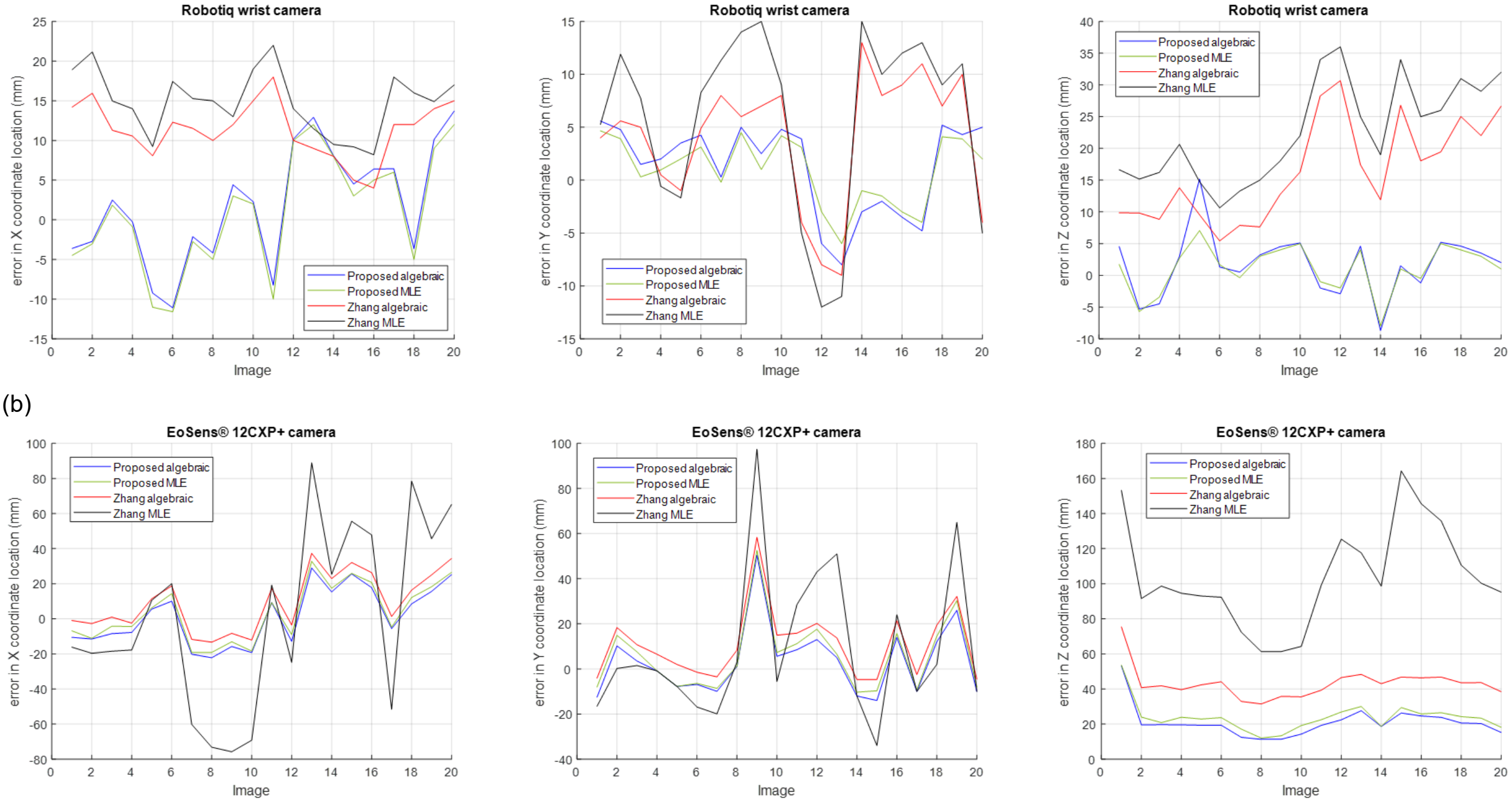

Fig. 15. (a). Error in locations for the Robotiq wrist camera. (b) Error in locations for the EoSens® 12CXP+ camera. In both cameras, the proposed method computes camera locations closer to the real one than the method proposed by Zhang. Analysis of the Zhang solution shows that, although the reprojection error is bigger, the algebraic result is closer to the real location than the MLE.

models with several scale factors using the method described in subsection III.B. An algebraic solution is computed with equation (20) that is improved with a nonlinear minimization problem defined in (21) or in (24) that includes the camera lens distortion parameters.

The aim is to minimize the reprojection error that represents the distance between the detected points in the image and the chessboard template points projected with the computed model. The Zhang method tries to minimize this index by modifying all camera parameters at the same time. With the proposed method, the scale factors $\alpha$ and $\beta$ of the matrix $\boldsymbol{A}$ remain constant to avoid the coupling between intrinsic and extrinsic camera parameters. The searched parameters are the extrinsic ($\boldsymbol{R}_i$ and $\boldsymbol{t}_i$) and the intrinsic ones (the principal point ($u_0$, $v_0$), skewness $\gamma$ and distortion, represented by $k_1$, $k_2$).

The mean values and the standard deviation of the reprojection error, with different solutions, are summarised in table 1. Table 2 shows the computed camera parameters with the proposed method and the Zhang method, including the algebraic and the MLE results.

Figure 15 compares several solutions of the position vector $\boldsymbol{t}_i$ computed with both methods and both cameras. Since camera true location is given by the robot arm, bias errors are obtained using the true camera location as the point of reference. The blue line corresponds to the bias with the algebraic solution computed with the method proposed in subsection 3.2. The green line corresponds to the MLE solution, computing only the subset of parameters proposed in this paper and assuming a constant focal length. The red line corresponds to the algebraic solution proposed by Zhang, and the black line corresponds to Zhang's MLE solution. Figure 15(a) corresponds to the errors in the location computed with the Robotiq wrist camera and figure 15(b) shows errors in location with the the EoSens® 12CXP+ camera.

TABLE 1

MEAN VALUES AND STANDARD DEVIATION OF THE REPROJECTION ERROR WITH DIFFERENT SOLUTIONS

| | Method | Median | Standard deviation |
|---|---|---|---|
| Robotiq wrist camera | Proposed method | 0.3362 | 2.1709 |
| | MLE Proposed method | 9.0165× e-11 | 1.5626 |
| | Zhang algebraic solution | 0.0472 | 1.8667 |
| | Zhang MLE solution | 1.8347× e-10 | 1.0487 |
| EoSens® 12CXP+ | Proposed method | -2.4143 | 14.4151 |
| | MLE Proposed method | -1.5807×e-10 | 3.1542 |
| | Zhang algebraic solution | 0.4644 | 9.0036 |
| | Zhang MLE solution | 1.2092×e-09 | 2.3201 |

*The reprojection error with MLE is smaller because parameters are adjusted to the data accurately. However, the differences between reprojection error varying all parameters and varying extrinsic parameters, distortion and skewness only is not significant.*

## 4.3 Discussion

Bias in the position vector $\boldsymbol{t}_i$ will give the accuracy of the computed camera parameters with each method. Analysing position vector $\boldsymbol{t}_i$ bias in figure 15, with both cameras, the camera positions computed with the proposed method are closer to the true positions than the positions computed

TABLE 2

CAMERA PARAMETERS COMPUTED USING THE PROPOSED METHOD IN ZONE 2 AND USING THE ZHANG METHOD

| | Param. | Proposed method | Zhang algebraic solution | Zhang MLE solution |
|---|---|---|---|---|
| Robotiq | $\alpha$ | 1370,8 | 1362,8 | 1319,7 |
| wrist | $\beta$ | 1373,8 | 1369,9 | 1329,7 |
| camera | $u_0$ | 645,8 | 651,7 | 711,2 |
| | $v_0$ | 359,3 | 370,5 | 375,9 |
| | $\gamma$ | 0,0001 | -0,002 | 0,0008 |
| | $k_1$ | 0.0087 | 0.0048 | -0.016 |
| | $k_2$ | -0.072 | -0.056 | -0.003 |
| EoSens® | $\alpha$ | 4755,8 | 4487,0 | 4289,8 |
| 12CXP+ | $\beta$ | 4789,1 | 4442,8 | 4295,1 |
| | $u_0$ | 2052,3 | 2046,8 | 2127,5 |
| | $v_0$ | 1548,5 | 1448,1 | 1566,7 |
| | $\gamma$ | 0,0012 | 0,0239 | 0,0280 |
| | $k_1$ | 0.0022 | 0.0421 | -0.1703 |
| | $k_2$ | 0.0570 | -0.134 | 0.1389 |

by Zhang method. With the proposed method, MLE adjusts the computed model to the input data, changing only the extrinsic camera parameters, distortion, skewness and principal point, and assuming that scale factors are accurate as they have been computed by specific trials. MLE does not significantly modify the result solution. However, in the case of Zhang's result solution, the difference between the algebraic solution and MLE is significant. Analysis of the Zhang solution shows that the algebraic result is closer to the true location than the MLE. MLE obtains worse results than the algebraic solution. Even though the reprojection error is bigger with algebraic methods, estimated locations are closer to the real ones. This is explained with the coupling between all camera parameters. The algebraic solution computes the camera parameters separately and the MLE solution tries to find out the best solution for all camera parameters at the same time. Solving the nonlinear minimization problem based on reprojection error by computing all the camera parameters is not a very good practice.

Analysing table 1, reprojection error values with MLE are smaller because the parameters are accurately adjusted to the data. However, the differences between the reprojection error computing all the parameters at the same time and computing all parameters assuming a constant focal length are not significant. This is because the parameters are adjusted to data but in one case all parameters change in every iteration of the optimization process and in the other case the focal length remains constant during the optimization process. A constant focal length sets a reference in the nonlinear optimization process that helps to reach an unbiased solution. Since all the parameters are tightly coupled, MLE is able to finish with a solution where the parameters satisfy the model for the given data but different of the true one. Indeed, the computed solution is represented by an optical axis in the Sophia plane which does not correspond to the true one. With the proposed method, intrinsic parameters, such as scale factors are estimated with specific tests, and MLE is used to compute extrinsic parameters, distortion, skewness and the principal point. Camera parameters are divided into two groups to avoid incorrect solutions due to the coupling between them. The constant focal length sets the reference to compute the unbiased optical axis in the Sophia plane.

As can be seen in table 2, with the Robotiq wrist camera, the algebraic solution that results from the Zhang method is closer to the one computed with the lens focusing in the hypefocal distance. However, the MLE solution is notably different. Analysing the results from the EoSens® 12CXP+ camera, the algebraic solution is different from that of the lens focusing in the hypefocal distance, and the MLE solution even more so. To explain these results, it is necessary to observe the range of distances that cover zones 1, 2 and 3, respectively, with each camera. Zone 2 of the Robotiq wrist camera, in which focal length does not change, is delimited by distances of 150mm and 1200mm. Capturing images of the chessboard closer to 150mm to the camera is not possible if the chessboard is printed in an A4 sheet. Consequently, since all images are taken in Zone 2, the focal length is constant and the algebraic solution of the Zhang method is similar to that of the lens focusing in the hypefocal distance, supposed as unbiased. However, when the algebraic solution is improved with the maximum likelihoold estimation, the tightly coupling between intrinsic and extrinsic parameters means that, although the final solution has a very small reprojection error, the computed parameters are biased.

Going deeper, if the chessboard were smaller than an A4 sheet because the application needs images closer to the camera, most of the calibration images would be in the zone 1 and camera parameters would be different to the ones computed with the lens focusing in the hypefocal distance. Moreover, if camera parameters computed with the Zhang method were used in a camera-object distance closer than 150 mm, results would be biased.

Analysing distances in Zones 1, 2 and 3 for the EoSens® 12CXP+ camera, Zone 2 (where the focal length is constant), starts at 700 mm and ends at approximately 2700 mm. With this camera, it is possible to put the A4 sheet calibration template in the image completely when the camera is 300 mm away from the calibration template. In consequence, images that are taken within a range of 300 mm and 700 mm are in Zone 1, where the focal length varies in order to obtain a sharp image. Therefore, using these images with the Zhang method disrupts the results because they were taken with a different focal length. The algebraic solution from the Zhang method does not compute good parameters similar to the ones computed when the lens is focusing in the hypefocal distance. Similar to the computed results with the Robotiq wrist camera, the MLE obtains an even more biased solution. When starting the MLE with biased parameters, the MLE will end producing more biased results due to the coupling between intrinsic and extrinsic camera parameters.

As the experiments show, the proposed method gives a stable solution to obtain an accurate camera calibration. The method proposed by Zhang could lead to incorrect results because of the camera parameters coupling and as images are taken at any distance and it does not define which set of images obtain the best solution. In most camera calibration papers an analysis of the number of images is performed, but the camera location is not defined.

## 5 Conclusion

In this paper, the influence of the focus in the calibration process has been analysed in depth. To obtain the sharpest objects in the image, lenses are adjusted depending on the distance of the camera to the object varying the focal length in each image. Since the pin-hole model does not consider this variation, computing the camera model using images captured from different distances to the calibration template could obtain incorrect results. Moreover, an exhaustive study of the interdependence between intrinsic and extrinsic parameters has demonstrated that they are tightly coupled. The focal length of the camera is linked to the distance of the camera to the object and the principal point is linked to the location X and Y of the camera in the scene. In consequence, computing all the camera parameters together in an iterative nonlinear minimization problem cannot be considered a good practice in general. It gives a valid result based on the reprojection error but the result does not represent the real camera.

The camera calibration process has been improved in three aspects. First, a camera model is proposed where several values of scale factors are used depending on the distance of the camera to the object in the scene. Secondly, the scale factors are computed with a specific, separate experiment, to avoid incorrect results due to the interference between intrinsic and extrinsic camera parameters. Thirdly, a nonlinear minimization problem is solved by computing only extrinsic parameters, the principal point, skewness and lens distortion instead of computing all camera parameters at the same time.

The proposed improved method accounts for the existence of several set points of the camera lens to obtain sharp images. In consequence, it is necessary a camera model with several values of scale factors to represent camera behaviour accurately. If a model with a constant value of scale factor is used, it is necessary to know the range of distances between the camera and the objects where these scale factors camera parameters are valid or conversely, to capture images of the calibration template in a range of distances valid for the application. In addition, to avoid the coupling effects between intrinsic and extrinsic camera parameters, it is advisable to conduct a specific test to separately compute each camera parameter instead of trying to solve the nonlinear minimization problem by computing all parameters at the same time. Computing all camera parameters at the same time in a nonlinear minimization process computes biased results. An improved method to compute the camera principal point with more accuracy, could be considered in a future work. At present, the proposed method is a step forward in the field of camera calibration that will help in any application where the camera parameters represent a crucial step.

**Carlos Ricolfe-Viala** was born in Valencia, Spain in 1973. She received the M. Sc. Degree in Industrial Electronics and Automatic Control Engineering in 2000 and his Ph.D. degree on Computer Vision, Robotics and Automatic Control in 2006 from the Polythechic University of Valencia (UPV), Spain. He is lecturer in the Polythechic University of Valencia since 2001 at the Systems Engineering and Control Department. He has participated in several national and international research projects. His research interests are image processing and computer vision, especially in motion estimation, feature detection and matching, camera calibration, 3D computer vision and intelligent system robot. He cooperates with the Institute of Industrial Informatics and Automatic Control of Valencia.